# A New Tool for Early Design Window View Satisfaction Evaluation in Residential Buildings

Jaeha Kim [a*], Michael Kent [b], Katharina Kral [a], Timur Dogan [a]

[a] Cornell University, Ithaca, 14853, N.Y., U.S.A.
[b] Berkeley Education Alliance for Research in Singapore, Singapore

*Corresponding Author[*]: jk2894@cornell.edu,*

*Environmental Systems Lab, Cornell University, Ithaca, 14853, N.Y., U.S.A.*

*Telephone: (+607) 255-5236*

**Abstract**

People spend approximately 90% of their lives indoors, and thus arguably, the indoor space design can significantly influence occupant well-being. Adequate views to the outside are one of the most cited indoor qualities related to occupant well-being. Analyzing view is becoming increasingly important as spaces with vistas and views to the outside are becoming a rare commodity due to urbanization and densification trends. However, to better understand occupant view satisfaction and provide reliable design feedback to architects, existing view satisfaction data must be expanded to capture a wider variety of view scenarios and occupants. Most research on occupant view satisfaction remains challenging in practice due to a lack of easy-to-use early-design analysis tools. However, early assessment of view can be advantageous as design decisions in early design, such as building orientation, floor plan layout, and façade design, can improve the view quality. This paper, hence, presents results from a 181 participant view satisfaction survey with 590 window views. The survey data is used to train a tree-regression model to predict view satisfaction. The prediction performance was compared to an existing view assessment framework through case studies. The result showed that the new prediction (RMSE=0.65) is more accurate to the surveyed result than the framework (RMSE=3.78). Further, the prediction performance was generally high ($R^2 \geq 0.64$) for most responses, verifying the reliability. To facilitate view analysis in early design, this paper describes integrating the satisfaction prediction model and a ray-casting tool to compute view parameters in the CAD environment.

# 1. Introduction

## 1.1 Importance of view quality

In the US, people spend more than 90% of their time indoors [1], and hence it can be argued that the design of our indoor environments has a significant impact on our physical health and psychological well-being [2]. The quality of indoor environments is determined by several parameters, such as thermal comfort, air quality, acoustics, daylight, and view quality [3]. Despite its importance as one of the indoor environment parameters, the view was seldom measured and has been overlooked within past indoor environment research [4]. According to the existing research on view quality's importance, high-quality views have also been related to quantifiable work performance, public health, and financial advantages in work environments. One study found that a sizeable peripheral window view containing vegetation, when looking at a desktop, improved office work performance (i.e., memory recall and mental function) by 10-25% compared with no view [5]. Another study found that in hospitals, in-patients who had access to a window with a view of a natural setting consistently recovered faster than patients that stayed in rooms with a view of a brick wall [6]. In addition, view quality can also affect rent prices by approximately 6% [7], and therefore adds considerable financial value. This research shows the importance of view quality and the necessity of view quality assessment.

Currently, consideration of window view (WV) can be seen in practice within rating systems (e.g. LEED v4.1 (EQ credit: Quality views)) and building standards (e.g. EN 17037: Daylight in Buildings [8]). The EN 17037 provides design recommendations by specifying thresholds for horizontal sight angle, the distance of view content, and the number of visible layers (i.e., foreground, landscape, and sky [9]). While the specification of design criteria in building standards encourages WV evaluation in buildings, most approaches generally rely on minimum acceptable thresholds to dictate view quality to avoid adverse effects and not necessarily promote occupant health and comfort [10].

The existing body of research on WV mainly focuses on the non-residential sectors such as schools, hospitals, and office spaces and relates view quality with employee work performance as well as physical and psychological comfort. Waczynska. *et al.* [11] presented a comprehensive literature review on EN 17307: Daylight in Buildings [8] view criteria, whereby only one of 26 cited publications focused on view in the residential setting. Since, with the emergence of SARS-CoV-2 and the lockdowns and work-from-home policies, people started spending most of their time at their own residence, which has changed the perception of WVs in residential spaces. During lockdowns, many people rearranged their room furniture in favor of a window-centered layout by relocating their desks closer to windows for daylight and view [12]. This was likely driven by the occupants' desire for a connection to the outside to observe weather, time, and location [13]. With this changed residential space usage pattern, there is a growing need to address WV quality in residential buildings.

**1.2 Quantifying view quality**

Many researchers investigated which variables can be used to determine view quality [4, 5] by examining different facets of the content that can be seen (e.g. water, sky, people, cars, and landmarks [4, 13, 14]). Other than visual aspects, factors dependent upon time, for example, within a relatively short period (e.g. weather and season), or across much larger time scales (e.g. building age, architectural style, and maintenance) can evoke either positive or negative responses when observed [15, 16]. Keighley *et al.* [18], and Abd-Alhamid *et al.* [19] found that window shape, the distance between view objects and observers, and visibility of certain visual layers (e.g. sky) influence view quality. Table 1 summarizes all significant view variables that have been identified in the literature.

To consolidate scientific literature and design standards, Ko *et al.* [20] proposed a View Quality Index (VQI) framework, constituting three primary variables: view content, access, and clarity. Multiplication of these variables (i.e. $\text{VQI} = V_{content} \cdot V_{access} \cdot V_{clarity}$) leads to an overall evaluation of view quality. This formula provides a relatively straightforward prediction method, but it does not consider the nonlinear relationships conventionally found between visual satisfaction and view quality [20, 21]. Since this formula is not verified by subjective appraisals (e.g. survey or experimental data), there remains a paucity of evidence on view satisfaction. Through a similar vein of research, Aries *et al.* [22] investigated demographic information (e.g. gender and age), circannual mood shifts, and the physical and psychological discomfort perceived by 333 office occupants. Their study showed that it is possible to estimate occupant levels of fatigue and satisfaction from the environment using demographic information and WV variables such as view type, view quality, and distance to the window. Nevertheless, the examination is limited to a specific office environment, and therefore the results are not easily transferable to a different space and building use, which makes it difficult to use the findings for general design evaluation.

**Table 1** Review of previous research on WV variables

| Variable(s) | Reference | Year | Experiment Setup | N | Evaluation Value |
|---|---|---|---|---|---|
| Window size / Window to wall ratio | Keighley et al. [18] | 1973 | Physical office model with projected view | 30 | Preference |
| | Ludlow [15] | 1976 | Physical office model with projected view | 20 | Preference |
| | Butler et al. [13] | 1989 | Questionnaire | 59 | Preference |
| View angle | Heschong Group [5] | 2003 | Office | 300 | Work performance |
| | Magistrale et al. [23] | 2014 | Projected windows with a view on a wall | 25 | Preference |
| | Moscoso et al. [24] | 2020 | VR | 406 | Feeling |
| Window width | Matusiak et al. [25] | 2016 | Participants' office windows | 106 | Preference |
| Window height | Ne'eman et al. [26] | 1970 | Physical classroom model with real window views | 318 | Acceptable window width |
| Aspect ratio | Ludlow [15] | 1976 | Physical office model with projected view | 20 | Preference |
| A number of windows | E. Ne'eman et al. [26] | 1970 | Physical classroom model with real window views | 318 | Acceptable window width |
| | Butler et al. [13] | 1989 | Questionnaire | 59 | Preference |
| | Magistrale et al. [23] | 2014 | Projected windows with a view on a wall | 25 | Preference |
| Window location | Keighley et al. [18] | 1973 | Physical office model with projected view | 30 | Preference |
| View angle | Heschong Group [5] | 2003 | Office | 300 | Work performance |
| Distance to window | Heschong Group [5] | 2003 | Office | 300 | Work performance |
| | Aries et al. [22] | 2010 | Participants' office windows | 333 | Physical & psychological discomfort |
| Room volume | Moscoso et al. [24] | 2020 | VR | 406 | Feeling |
| Sky condition / Seasonality | Kaplan [27] | 2001 | Image Questionnaire | 188 | Preference |
| | Aries et al. [22] | 2010 | Participants' office windows | 333 | Physical & psychological discomfort |
| | Matusiak et al. [25] | 2016 | Participants' office windows | 106 | Preference |
| | Moscoso et al. [24] | 2020 | VR | 406 | Feeling |
| View of nature and building | Ulrich et al. [14] | 1981 | Patient's rooms | 46 | Length of hospitalization, patient's condition, analgesic does amount |
| | Ulrich et al. [6] | 1984 | Image slide | 18 | Feeling, reaction, heart rate, and alpha altitude |
| | Kaplan [27] | 2001 | Image Questionnaire | 188 | Preference |
| | Aries et al. [22] | 2010 | Participants' office windows | 333 | Physical & psychological discomfort |
| | Hellinga et al. [28] | 2014 | Image Questionnaire | - | View quality |
| | Matusiak et al. [25] | 2016 | Participants' office windows | 106 | Preference |
| View of water | Ulrich et al. [6] | 1984 | Image slide | 18 | Feeling, reaction, heart rate, and alpha altitude |
| | Hellinga et al. [28] | 2014 | Image Questionnaire | - | View quality |
| | Matusiak et al. [25] | 2016 | Participants' office windows | 106 | Preference |
| View of dynamic | Kaplan [27] | 2001 | Image Questionnaire | 188 | Preference |
| | Hellinga et al. [28] | 2014 | Image Questionnaire | - | View quality |
| Building style and maintenance | Hellinga et al. [28] | 2014 | Image Questionnaire | - | View quality |
| Number of view layers | Hellinga et al. [28] | 2014 | Image Questionnaire | - | View quality |
| | Matusiak et al. [25] | 2016 | Participants' office windows | 106 | Preference |
| Distance to view objects | Ne'eman et al. [26] | 1970 | Physical classroom model with real window views | 318 | Acceptable window width |
| | Matusiak et al. [25] | 2016 | Participants' office windows | 106 | Preference |
| Landscape distance | Kent and Schiavon [29] | 2020 | Online Image + test room with screen | 141 | Satisfaction of connection to outside, content, privacy |
| Space program | Moscoso et al. [24] | 2020 | VR | 406 | Feeling |

## 1.3 Tools to evaluate view quality in practice

Despite the growing availability of design and environmental simulation tools, there remains a need for software that can conveniently evaluate and optimize view quality in early design. View quality standards are currently defined with simple checks such as view angles that can be measured in 2D plans and sections, allowing designers to compute the proposed view quality metrics manually. Nevertheless, this kind of evaluation is still cumbersome, especially for a larger building with many rooms, and therefore view quality assessment remains impractical during the design process. In computer-aided design (CAD), view content can be quantified computationally by testing a 3D model for visibility using rays – this approach is known as Ray-casting and was first introduced by Benedikt [30]. Hwang and Lee [31] used Ray-casting from a given point in a room to measure window parameters and obtain the 3D visible area from the observer's position. The ray intersection is used to identify the view element type (e.g. building or trees) as seen from the viewing position of an occupant looking through a window inside a scaled 3D site model. Based on this ray-casting method, Doraiswamy *et al.* [32] and Ferreira *et al.* [33] suggested a view analysis framework and proposed a catalog-based design evaluation tool. In their framework, view quality is decided by the depth of view, visibility of landmarks and the landscape, and variation of visible building styles. The framework is developed as a view evaluation software and is used for design discussions with clients, considering their preferences. This framework is further annexed to an urban scale data-driven decision-making pipeline Urbane, which has been applied to the Manhattan design-development case study [33]. On the other hand, the popular environmental performance analysis software Climate Studio [34], provides EN 17037 and LEED view index auto-computing functions for designers and engineers. However, no existing tool can predict occupant view satisfaction using a data-driven model (i.e. based on view surveys) that can interpret different view variables and their complex interaction with occupant satisfaction.

## 1.4 Literature gaps and motivation of this work

***Small training data sets limit existing occupant view satisfaction predictions:***

Considering the importance and need for view evaluation, software that can conveniently and holistically evaluate view quality for occupant view satisfaction in the design process is needed. Existing research surveys and experiments on view satisfaction often use a limited number of view samples. Among 26 selected studies by Waczynska. *et al.* [11] on view assessment, 23 studies used an average of five view scenarios, while only three studies used an average of 112 view scenarios. Due to the small number of views used in most studies, past findings may not capture every nuance for each view, and therefore results from statistical prediction models cannot easily be generalized. To diversify the view scenarios, the authors conducted an online survey, asking 181 participants to rate 590 unique WV configurations of residential settings. The intent is to enhance the fidelity of view satisfaction analysis by covering a wider range of view scenarios.

***3D model preparation for view evaluation is cumbersome and therefore rarely used in design:***

Occupant view satisfaction assessment remains a challenging task in design practice. When evaluation frameworks are conceived, they often remain theoretical and are not implemented as ready-to-use tools that are integrated seamlessly into design processes. In addition, the configuration and setup of a 3D model that can be used for view evaluation is often a manual task that involves sorting geometry into layers to figure out what kind of view content category they represent. To facilitate occupant view satisfaction evaluation in design practice, the new tool presented in this paper has been integrated into the popular Rhino/Grasshopper CAD ecosystem.

This implementation can further load labeled contextual geometry from GIS and OpenStreetMaps [35] through Urbano.io [36] to significantly facilitate the 3D model setup.

***Open-source survey and training data to facilitate continuous improvement of the* occupant view satisfaction prediction *model:***

The training data and survey used in this research are published as an open-source repository to facilitate future research and continual improvement of the presented occupant view satisfaction prediction model. The repository recognizes the need to expand the number of view scenarios that the model has been trained on as well as to diversify the survey respondents to obtain a more representative satisfaction rating. Even though this research endeavored to predict window view satisfaction using the largest known sample of views, it is important to establish the foundations for a publicly available dataset to continually advance the knowledge in this domain.

## 2. Methodology

The research methodology consists of three main steps. First, the authors conducted an online survey that asked participants to rate view satisfaction on 590 different WV samples. Second, the authors developed a computational tool that uses ray-casting to extract key view properties, including view parameters from a 3D CAD model for different view content categories and view composition parameters. The tool is called "Seemo-Raycaster" and is implemented in C# and integrated into the Rhino [34] design environment. Third, the authors proposed a new WV satisfaction evaluation framework based on Ko *et al.*'s [20] primary variables, leveraging a machine-learning (ML) model to synthesize data for 590 WV samples from both survey responses and 3D modeled view parameters.

The trained view satisfaction prediction engine – herein, "Seemo-Predictor" – is also integrated in Rhino and can make view satisfaction predictions for any given 3D architectural model. Lastly, to demonstrate the efficacy of the tool's high fidelity, the authors tested the framework in two case studies and compare the prediction results with results obtained from the survey and previous work on view quality by Ko *et al*. [20].

### 2.1 Online survey design

The survey was designed to collect subjective evaluation data on view content, view access, privacy, and overall satisfaction. In total, 590 unique view scenarios were prepared for evaluation. These unique view scenarios are composed out of 59 views and are then overlaid with ten different window configurations. The authors carefully selected the 59 photos showing views with different visual elements (e.g. building, nature, people, infrastructure, etc.), which had content at varying distances, number of layers (e.g. landscape and sky), and were viewed from different floor elevations. The authors collected images that were plausible views occupants may encounter in urban areas. The photos were mostly taken by the authors, while the rest were either purchased from Adobe stock photos or were sourced from previous studies [28, 36]. The latter were used to facilitate comparisons to previous findings. For each of the 59 view images, ten window types were evaluated (Figure 1). The authors chose window types with different sizes, aspect ratios, subdivisions, and window-to-wall ratio. Nominal room size was set to 3.8m (width), 4.3m (depth), and 3m (height). The camera was positioned in the center of the room at 1.2m height, and the image was produced with a commonly used 3:2 view frame ratio [40].

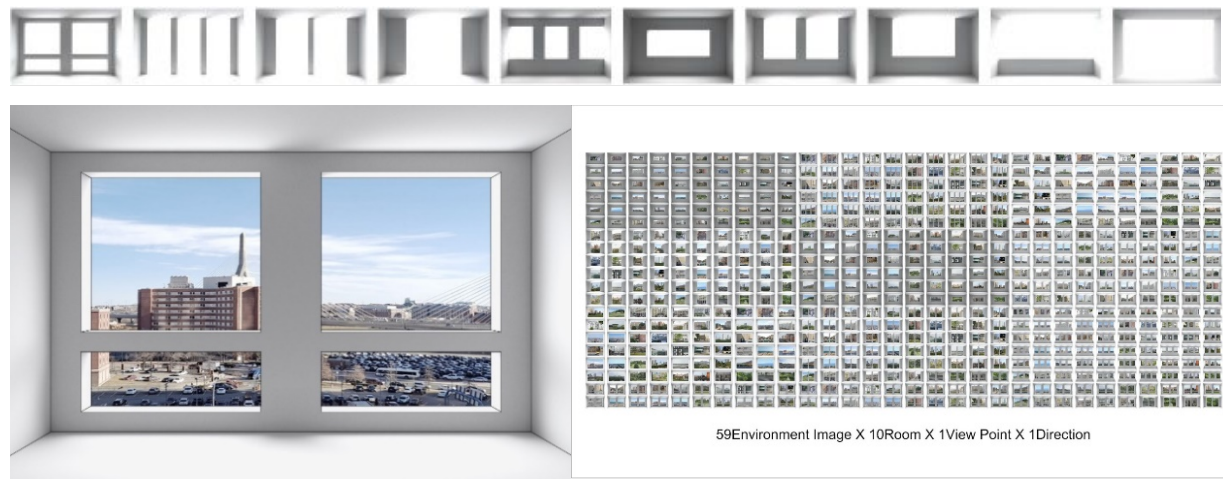

**Figure 1** The ten window types and 590 image samples used to evaluate view satisfaction

Instructions were presented at the beginning of the survey, requiring respondents to provide demographic information (i.e. age, gender, origin, architectural background, eyewear, and eye health). A collage of every view was presented at the start of the survey, allowing participants the opportunity to see the full range of images they were about to evaluate [41]. For each image sample, participants rated their degree of visual satisfaction on four criteria: view content [9], view access [17, 22], privacy [9], and overall rating [42]. Participants were asked to imagine each image was their living room view and gave their ratings to each criterion based on the following descriptions:

- View content: Refers to what you see in the view (e.g. buildings, people, and/or sky).
- View access: Evaluates the adequacy of the window size and distance from the window.
- Privacy: Evaluates concerns on privacy due to view content and window configuration.
- Overall rating: A general evaluation for the view and window that you see.

Ratings were given on continuous scales ranging from -5 (unsatisfactory) to +5 (satisfactory) with the slider set to the neutral point (0) by default. To reduce ordering effects, survey images were randomized [43]. Participants were randomly allocated 30 of the 590 available images to avoid fatigue [44].

The survey was distributed to first-year architecture students and their instructors and received 181 responses. Eighty-four participants were male and 85 were female, four were aged under 18, 107 were aged between 18-24 years of age, 46 were between 25-34 years, six participants between 35-44 years, and three participants between 45-54 years. One hundred and twenty-seven participants already had an educational background in architecture, and 36 did not.

## 2.2 View property computation

The Seemo-Raycaster computes view parameters (Table 2) to visual elements found within a given 3D model, rather than simply measuring views' general composition (e.g. horizontal layers) as often done with image-based view analysis [24, 42]. To compute these view parameters, rays are cast through a view plane into the 3D model that is positioned in front of a user-defined view position and direction. The user can also specify a resolution that determines the x, and y resolution of the view plane (Figure 2a). A finer resolution tends to increase view parameter accuracy and can detect small scene objects but comes at the cost of increased computational overhead. Image resolution was consistently set to 366px×244px. The ray-casting process is divided into two steps. First, the room model including interior walls, ceiling, facades, and window geometry is used to determine which rays hit a window or opening. Based on this first step, the window view parameter can be computed simply by counting window rays and dividing them by the total number of rays sent out. In the second step, rays that intersected with a window, are then cast into the contextual 3D model to determine the view parameters of each view content category (e.g. buildings, trees, etc.) that is represented in the 3D model (Figure 2b).

To measure a perceived distance for each element type (Figure 2b) [19], the tool calculates the average distance of the closest 30% of ray-object intersection points for each view category. This method which calculates a critical distance to several objects in the same content type is devised to avoid a distance to an unrecognizable minuscule object but to capture a close average distance instead of middle distance or far distance because perceived view depth is mostly influenced by close objects obstructing the view. To incorporate aspects of view composition, the tool uses the "Rule of Third" [46], to divide up the view plane into four rectangular regions (Figure 2c): Middle and center (Z1), Middle and Side (Z2), Top (Z3), and Bottom (Z4). For each region, a separate window view parameter is reported. Since overall WV changes with the floor level of a room [39], this is calculated by measuring the height difference of the view location and the ground surfaces in the 3D model. This parameter can also be user input. Further, changes in weather were accounted for but relies on user input since this information is rarely included in the architectural 3D models. The tool encodes the sky condition according to: invisible sky= 0; overcast= 1; and clear sky= 2.

**Table 2** List of the 23 WV variables computed by the Seemo-Raycaster

| Category | Variable | Category | Variable |
| --- | --- | --- | --- |
| Window Size (2) | Wn: Number of windows<br>Was: Sum of window areas | View parameter (8) | Br: Building View parameter<br>Er: Equipment View parameter<br>Tr: Tree View parameter |
| View Composition (4) | Z1r: Middle and center located window area ratio<br>Z2r: Middle and side located window area ratio<br>Z3r: Top located window area ratio | | Pr: Artificial Ground View parameter<br>Gr: Ground Vegetation View parameter<br>Wr: Water View parameter |

|  | Z4r: Bottom located window area ratio |  | Dr: Dynamic View parameter<br>Sr: Sky View parameter |
|---|---|---|---|
| Additional Parameters (3) | EN: Number of Visible Element Types<br>*View diversity<br>FH: Floor Heights of the viewpoint<br>SC: Sky Condition<br>*Invisible sky(0), Overcast sky(1), Clear sky(2) | Distance to View Category Objects (6) | Bd: Distance to Building<br>Ed: Distance to Equipment<br>Td: Distance to Tree<br>Gd: Distance to Ground Vegetation<br>Wd: Distance to Water<br>Dd: Distance to Dynamic |

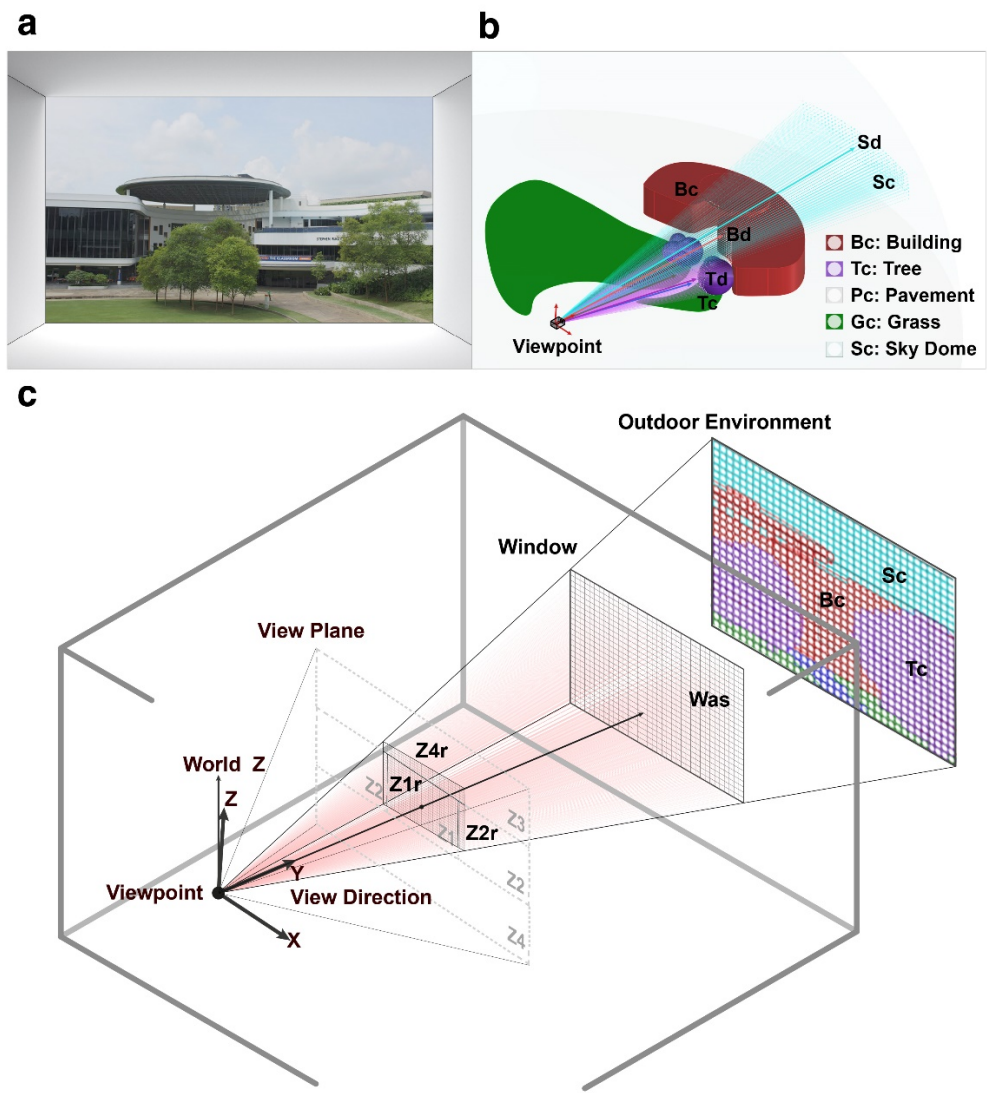

**Figure 2** Images showing the different stages of development: (a) fully unobstructed view plane, framed by the surrounding walls, floor, and ceiling; (b) Ray-Casting and View Object Detection; and (c) Ray-Casting and WV Variable Detection

## 2.3 Survey Image 3D Model Reconstruction

To compute view properties using the Seemo-Raycaster, a 3D model is needed. Hence, 3D models of all view scenarios used in the survey had to be reconstructed to obtain a consistent dataset for model training. For the 59 images used in the survey, 3D site models for each had to be created (Figure 3). The 10 facades were modeled individually and then combined with the site models to replicate the 590 view scenarios. For most images, the site and its surrounding context could be modeled from geotag information downloaded from OpenStreetMaps data [35] using Urbano.io [36] (Figure 3d). The perspective of the sample view was then reconstructed by iteratively moving the Rhino viewport camera until a good match (Figure 3b) with the photograph (Figure 3a) was achieved. Finally, details missing from the OpenStreetMaps data such as building details and trees (Figure 3c) were added manually to match the photo.

In some cases, however, the exact location of the photo was unknown and/or no 3D data was available on OpenStreetMaps. For these cases, a manual reconstruction approach was used. Scenery found in each image was modeled in a 5-step process: 1) A background image was placed on the wall of the 3D room model and scaled to the size of the window; 2) Two Rhino3D viewports were created: one parallel to the wall, and the other set to perspective (i.e. matching the camera setting used for window image rendering); 3) Within the parallel viewport, every object in the background image was traced along with object's silhouette; 4) The traced objects were scaled based on size estimation (e.g. one-story height was approximate to 3.2m). The origin point of the scale was set as the perspective camera location; and 5) The actual-sized objects are moved along the view direction to fit in the location on the perspective view.

All 3D geometry was categorized by their view content category (e.g. building, trees, dynamic) and organized using Rhino layers. For the sky layer, a sphere (radius: 50km or greater than the site model) was created. Its center point corresponded with the viewpoint.

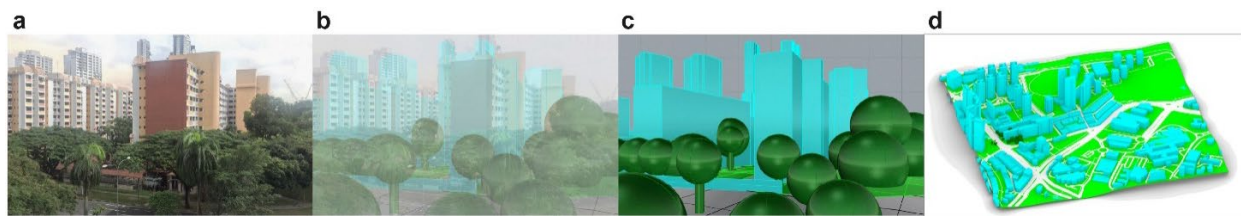

**Figure 3** Example of 3D model reconstruction of photo-based view scenarios: (a) a background image from a survey; (b) matching process; (c) environment detail edition; (d) 3D context model

## 2.4 View satisfaction prediction model "Seemo-Predictor"

Using the survey responses and view property features computed with the Seemo-Raycaster, supervised ML models were trained to independently predict survey responses to overall rating, view content, view access, and privacy. To train supervised ML models, these view satisfaction ratings (labels) and the view property features computed by the Seemo-Raycaster (features) needed to be combined and trimmed to generate one consistent dataset.

In the online survey described in section 2.1, 181 participants were asked to rate 30 WVs. Three percent of the collected data comprised of missing responses. To improve the reliability of ML model predictions by removing outliers in the data, two data trimming processes were applied when either one of the following conditions was met: 1) Responses were more than 1.5 interquartile range – either below the 25$^{th}$ percentile or above the 75$^{th}$ percentile [47][48]; and/or 2) Responses had a standard deviation greater than three units on the visual satisfaction scale. These two data trimming processes were conducted using STATA 17, which trimmed 11% of the dataset. In total, 4659 responses were used in ML model training. Individual view ratings were converted into a mean rating for each view scenario, following the procedure by Aries *et al.* [22]. The dataset was then split into model training and model validation data using an 8:2 ratio split.

ML model selection and training was done with Microsoft's ML.Net AutoML feature that automatically tests several regression models [49] and then selects regression models with the highest predictive performance. In this research, the authors trained AutoML for 10 seconds for each label, testing 40-50 models within the given period. Three tree-based models were considered: namely, Fast Forest Regression (random forest [50]), and Fast Tree Regression [51]. Tree-based models were expected to perform well on the dataset since they tend to work well with high-dimensional data and for continuous predictor variables [52].

Permutation Feature Importance (PFI) was used to interpret the relative contribution of each feature making a prediction [51]. These analyses revealed which computed view properties were the most relevant for the prediction of the four view satisfaction ratings (labels). The authors selected the four best performing models ($M_{y1}$, $M_{y2}$, $M_{y3}$, $M_{z1}$) from AutoML by including 23 view features into each model, and verified its performance based on changes in the coefficient of determination ($R^2$).

## 2.5 Case study 1

In the first case study, the predictive performance of the Seemo-Predictor was compared to Ko *et al.*'s [20] framework on the six WVs published in their work. Since the latter had only presented data for view content, comparisons could not be extended to the three variables. Calculations were performed following the approach recommended in the WV framework (Appendix A) [20]. To provide direct comparisons of Ko *et al.*'s framework and the Seemo-Predictor, the same viewing position in the surveyed WVs, view features, and other parameters were used (Appendix B). To enable statistical comparison, the framework's view quatitative values (ranging from zero to one) were normalized to match the Seemo-Predictor's satisfaction scale ranging from -5 to +5.

## 2.6 Case study 2

Four image pairs (i.e. eight WVs) were considered for further evaluation (Table 3) in the second case study. Each pairing juxtaposes images with overall similar view content and window, but with subtle differences: For example, the views in Pair 1 both have a fully glazed façade with views of distant high-rise buildings, and nature (i.e., high visual content score, view access score according to Ko *et al.*'s framework). The difference is that #4 includes vehicles, while the scene in #394 shows people and water. Both vehicles and people are considered as features that intrinsically are dynamic (i.e. they frequently and overtly move). Pair 2 both images include sky and buildings that are located at the same distance from the window. Both images also include people but the distance of the people in the scene varies: #550= 17m, and #520= 44m. Pair 3 both images show water and dynamic objects (boats and people) through fully glazed façades. However, sky conditions and dynamic object types in the two images are different. Pair 4 features fully glazed façades with views of nearby buildings and no ground or sky visibility (i.e., low view content, but high access). The major difference in Pair 4 is that the view in image #284 includes building HVAC equipment. These smaller nuances cannot be captured by the Ko *et al.* framework but are considered in the presented model due to the introduction of additional view content labels and high-fidelity measurement of the tool. The authors test that whether capturing these differences would improve the new WV evaluation framework compared with the existing Ko *et al.* framework.

**Table 3** Four image pairs (eight WVs) and their variable information, demonstrating similarities and differences within each scene

| Pair | Images | | Variable information |
|---|---|---|---|
| 1 | 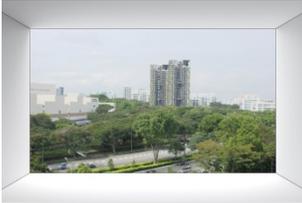 #4 | 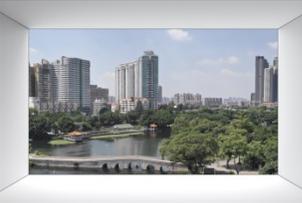 #394 | • Similarity: Buildings, nature, and clear sky<br>• Difference: Water, and dynamic object type (vehicles, and people) |
| 2 | 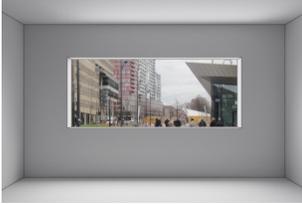 #550 | 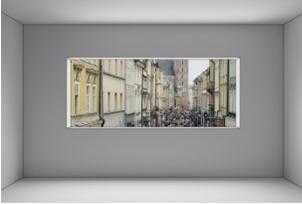 #520 | • Similarity: Buildings, people, and sky<br>• Difference: Distance to people |
| 3 | 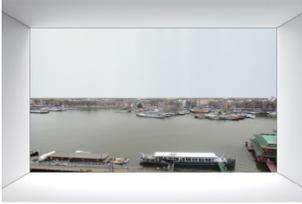 #334 | 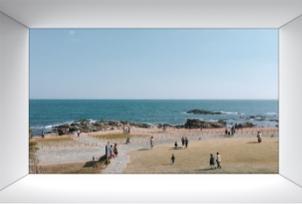 #294 | • Similarity: Water<br>• Difference: Sky condition, and dynamic (boats, people) |

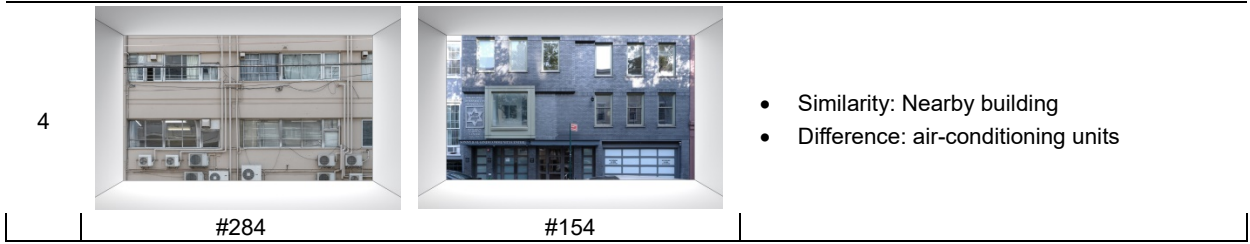

| 4 | #284 | #154 | • Similarity: Nearby building<br>• Difference: air-conditioning units |

## 3. Results
### 3.1. Model performance and Permutation Feature Importance

AutoML tested 40-50 models for 10 seconds. Table 4 shows the best performing models with the highest $R^2$ and lowest *RMS-loss*. $R^2$ of overall rating is the highest and is $R^2$ of view access is the lowest. With the exception of view access, ML models for the other three labels gave high predictive performances ($R^2 \geq 0.64$) [53]. *RMS-loss* is highest for privacy and is lowest for overall rating. Overall rating has the highest $R^2$ and lowest *RMS-loss* among four labels (i.e., this was the best overall performing model).

**Table 4** Top1 model performance for the four survey responses

| Ranking | Overall Rating $Z_1$ | View Content $Y_1$ | View Access $Y_2$ | Privacy $Y_3$ |
|---|---|---|---|---|
| 1st Model | Fast Tree Regression ($M_{z1}$) | Fast Tree Regression ($M_{y1}$) | Fast Tree Regression ($M_{y2}$) | Fast Forest Regression ($M_{y3}$) |
| $R^2$ | 0.90 | 0.89 | 0.61 | 0.76 |
| *RMS-loss* | 0.66 | 0.88 | 0.68 | 1.07 |

Figure 4 shows the results of the PFI analyses. To present how feature importance varies for the 23 features across the four parameters, a bump chart was produced. Features were initially ordered by their relative importance for the prediction of overall rating. Seven of the first 10 features for overall rating remain in the first 10 in privacy, five of the first 10 features remain in the first 10 in view content, and four of the first 10 overall rating features remain in the first 10 in view access. Considering that the variability in feature ranking was smaller when features transitioned to privacy and view content, those have a larger influence on overall ratings compared to view access. Features that consistently appear to be important across all four facets are: Sky view parameter and floor height of the viewpoint.

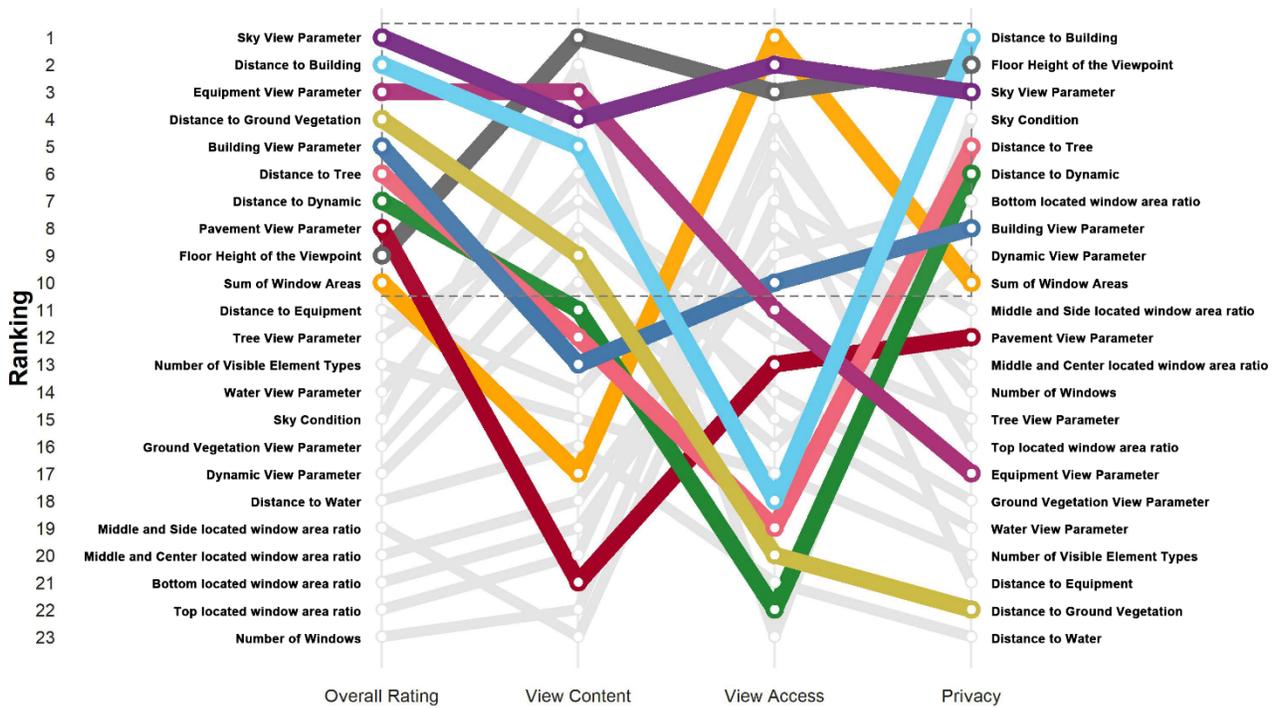

**Figure 4** Bump chart of PFI results for the four survey responses

### 3.2. Case Study 1 Results:

Figure 5 compares the view content satisfaction evaluation results between the Ko *et al.* framework and the new ML model using the six images that were used in previous research. The comparison result is 1.01 mean absolute error (MAE), and 1.42 root mean square error (RMSE). The error unit is the same as the continuous scale used to measure visual satisfaction. This demonstrates that the two sets of data generally show agreement in their prediction given to the six images. Nonetheless, discrepancies emerge in #234 and #424.

In #234, the difference is attributed to the calculation method for some of the larger view elements. The new ML model presents element type (e.g. sky, and buildings) area as a view parameter. However, Ko *et al.*'s framework evaluates these and similar elements by four inclusive visual layers: sky, landscape (e.g. building), ground, and nature. Each is assigned a weighting option, either zero (i.e. not present) or 0.25 (i.e. present) contributing to an assessment of view content and overall quality. For view #234, the sky is visible, but the area is small. The framework assigns a weight of 0.25 to the sky layer, but the Seemo-Predictor assigns the more precise value representative of its area (Sky View parameter = 0.03 (Appendix B)).

In #424, the difference is caused by the dynamic information assessment method. Although dynamic view content is in itself stochastic, both methods provide assessments for moving features. The Seemo-Predictor differentiates dynamic view elements (e.g. people or vehicles) from those that are static (e.g. buildings or trees) and computes dynamic view parameter and distance to dynamic element respectively, while in the Ko *et al.*'s framework movement (denoted "wf$_{movement}$") is determined both by the presence of movement, and the distance it can be seen from the view. This classification is expressed by a ternary operator: 1= distant movement, 0.5= no movement, and 0= nearby movement. As a result, the Seemo-Predictor evaluated the #424 dynamic view parameter as 0 and the distance to dynamic element as NA, however, the framework evaluated the #424 movement as 0.5.

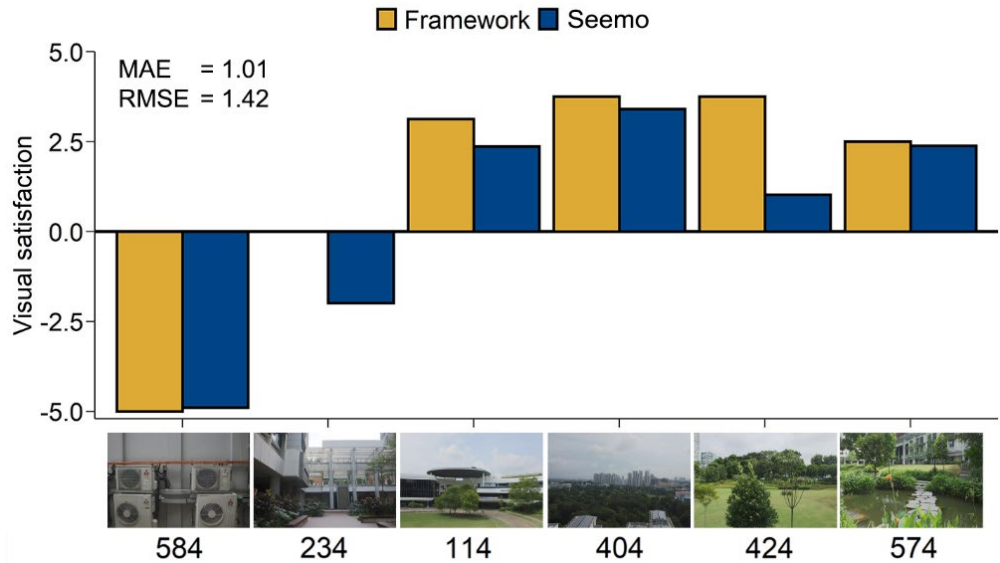

**Figure 5** Comparison between the Seemo-Predictor and Ko *et al.*'s framework for view content

### 3.3. Case Study 2 Results:

Figure 6 compares the evaluations of the Seemo-Predictor, Ko *et al.*'s framework, and the survey responses on the four image pairs (Table 3). The comparisons are made for: (a) overall rating, (b) view content, (c) view access, and (d) privacy. MAE and RMSE represent the differences between survey responses with Seemo-Predictor, and Ko *et al.*'s framework in each plot. The differences appear to be consistently larger between Ko *et al.*'s framework and the survey responses and were smaller when the latter is compared to the Seemo-Predictor. This indicates that the tool can provide predictions that are closer to actual responses given by the survey participants. The smallest differences between the tool and the survey are for (a) overall rating (MAE = 0.53, RMSE = 0.65), and are largest for (d) privacy (MAE = 1.31, RMSE=1.60).

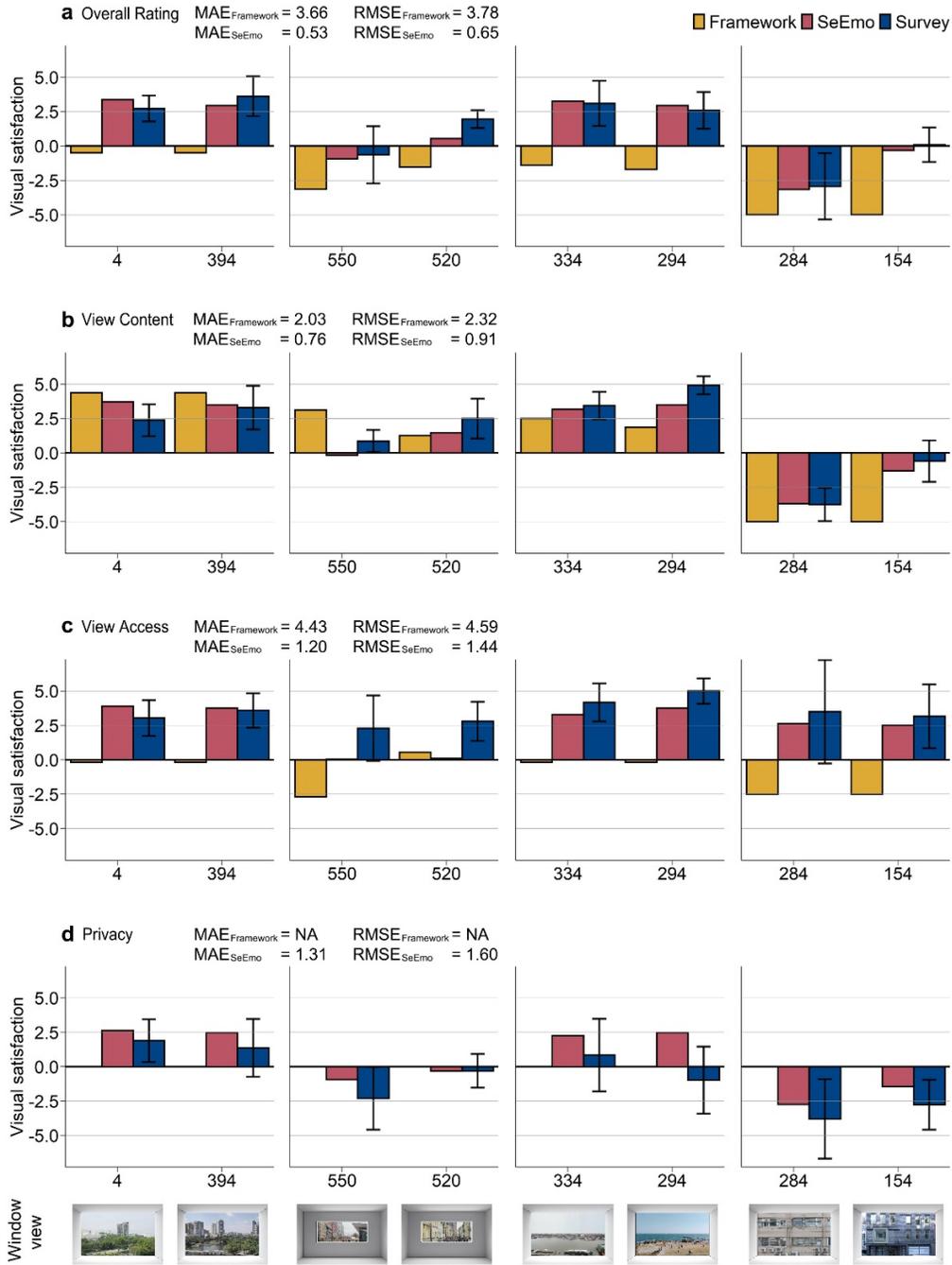

**Figure 6** Comparison between survey responses, Seemo-Predictor, and *Ko et al.'s* framework for (a) Overall Rating, (b) View Content, (c) View Access, and (d) Privacy.

When comparing the relative score differences from the survey responses with the Seemo-Predictor for all plots and pairs, only two cases showed discrepancies. In view content (Figure 6b), Pair 1 (#4 and #394 in Table 3) Seemo-Predictor does not correspond with the survey result. Two images look seemingly similar, except that #394 shows water and #4 shows the road. However, the Seemo-Raycaster evaluates two images in more detail. According to the evaluation result (Appendix C), #4 has less building view parameter, larger sky view parameter, and tree view parameter, which all contribute to view content. The tool captures that #394 shows water, but the water view parameter is 0.12, and the distance to water is 319.62m. Since the view parameter is not significant and distance is too far the tool might predict #4 is more satisfying than #394 in terms of view content.

For privacy (Figure 6d, images #334 and #294)), predictions given by the Seemo-Predictor do not accurately correspond with the survey responses. Although the underlying reasons for this are not entirely apparent, it could be due to the tool's inability to distinguish between boats and people (i.e. both features are classified as dynamic). Unwanted privacy caused by boats may not be perceived in an equal manner to similar issues produced by people.

## 4. Discussion

This paper introduces the Seemo-Raycaster which is capable of rapidly computing myriad WV variables. The novelty of this work is centered on the need for a high-fidelity tool that can reliably predict overall view quality and constitute facets (i.e. view content, access, and privacy), which can aid designers at the early design stage. This is also driven by the inert need to produce high-quality views that lead to high occupants' satisfaction and ultimately can support the health and well-being of residents.

### 4.1 Enhancing design sensitivity for residential occupant view satisfaction prediction

Through the online survey, the authors collected empirical test results on 590 window view samples. This overcame an overarching caveat for exant studies, which on average, analyzed the influence of a smaller number of variables on five different views [11]. Pitfalls from past research were avoided by using a larger variety of view samples, and WV variables postulated to have a strong influence the primary variables constituting window view [20]. Moreover, visible view object category labels are refined in this research. For example, other view analysis frameworks track a "building" view content category. We distinguish between "building façade" and "visible HVAC equipment". Further, the commonly used "nature" view category is broken into , "vegetation" and "water bodies". In case study 2, the framework did not capture the view quality difference of #284, and #154, while the Seemo-Predictor predicted satisfaction difference of the two images based on "equipment view parameter" In addition, sky conditions are labeled, and distance to view objects are precisely measured. As a result, the Seemo-Predictor is able to evaluate the degree of privacy invasion depending on the distance to dynamic objects, negative effects of equipment view parameter on view, impacts of sky view parameter on satisfaction. This enhanced sensitivity also enables accurate prediction of satisfaction on WV.

*Relative Importance of 23 variables in four labels*

A large number of image samples facilitates holistic analysis of 23 WV variables. The authors examine the relative importance of variables through PFI in section 3.1, revealing which variables are more critical than others. The PFI result provides useful information in design decision-making, since there could be trade-offs.

PFI shows that the sky view parameter is ranked higher than trees and vegetation for the prediction of overall view satisfaction rating. This is in stark contrast to several previous studies that focused on nature view content in window view studies, demonstrating that nature was the most important factor in view quality [13, 26, 56]. For instance, Ulrich *et al.*'s [14] compared natural scenes with urban scenes and concluded that these differences cause psychological effects. However, most of the natural scenes include the sky in the view, and the urban scenes normally do not include any sky. There could be collinearity between natural objects and sky. Comparison of #424 and #574 also implies a similar interpretation (Figure 5). Both images have the same amount of nature view parameter, #424 showing sky is more satisfactory than #574. The authors assume that this difference is caused different sky view parameters, according to PFI.

*Relationship between the overall rating, view content, view access, and privacy*

In addition, WV variables which are important to the prediction of overall ratings are almost equally important to the prediction of privacy and view content. This suggests that privacy and view content are highly related to the overall rating of a living room WV. Considering that designers may generally opt to promote overall view satisfaction, visible element types (e.g., sky, Equipment, building, pavement, ground vegetation) are among the variables that were important. Distance of certain variables (e.g. distance to buildings, ground vegetation, trees, dynamic) was also found to be among the strongest indicators for overall ratings. Besides overall quality, the utility of this tool can also leverage specific attributes of the view (e.g. content or privacy) to meet certain building or space program requirements.

**4.2 Examination of the new view satisfaction prediction tool**

The aforementioned findings were corroborated by the Fast Forest Regression model and the Fast Tree Regression model for the four surveyed labels (Table 4). WV features given by the tool were used in producing high prediction accuracy ($R^2 \geq 0.64$) [53] for overall view ratings, view content, and privacy.

The authors compared Seemo-Predictor with Ko *et al.*'s framework through the two case studies to test Seemo-Predictor. First, the authors compared predictions given to the six images published in the framework with predictions compuated by Seemo-Predictor. Both predictions gave relatively similar evaluations for view content (Figure 5). Small discrepancies emerged, but the differences were likely attributed to relative (i.e., scaled from zero to one) estimations given by the framework for view features, and precise (i.e., actual/weighted measurements) values given by the tool. Despite comparisons showed general agreement in their predictions, it was difficult to ascertain which was more accurate. Thus, predictions given to four view pairs by the framework, Seemo-Predictor, and survey responses were compared. The new tool consistently outperformed the framework (Figure 6). In other words, the tool's predictions were closer to survey responses. The authors assumed that the granularity and high fidelity on measuring features (e.g. water, grass, and Equipment) helped to minimize differences in predictions given by the tool and survey responses provided by participants. Although Ko *et al.*'s framework distinguished many differences across these image pairs, within each pair, it did not distinguish the more intricate differences of features (e.g. water, sky condition, or equipment element type (air-conditioning units)). As a result of the comparisons, the authors found that applying ML models and analyzing views using the high-fidelity Seemo-Raycaster tool improve the predicting performance.

### 4.3 Limitations and Future Study

*View access accuracy*

According to section 3.1, view access prediction accuracy was the lowest among the four labels ($R^2$<0.64). View access prediction may not provide equally reliable predictions for this aspect of the view. This result might be caused by the methodological constraints of this research (e.g., WV images). The survey samples may not be diverse enough to evaluate every facet of view access. Accessibility is determined by the occupant viewing distance and location [25], and not solely by the shape and size of window openings. In this study, views are always evaluated from an on-axis gaze position and at a constant viewing distance from the window, while these may vary more widely in-situ. Although the WV samples showed the edge of floor and ceiling, giving some sense of scale and depth, no other spatial cues (e.g. furniture) to enhance the sense spaciousness of the room were provided. A future experiment using VR could overcome the abovementioned issues by rendering or simulating realistic indoor conditions, and retaining controlled over the visual environment and parameters of experimental interest (e.g. the view).

However, it is unclear if a reliable assessment of view access is as important in relatively small residential spaces in comparison to work environment since view access is largely determined by the occupant's distance to the window. Further, in residential space, occupants have more flexibility to rearrange furniture in a better position and orientate their seatings to the window. Interestingly, many of the variables denoted important to the prediction for view content and privacy were not as important for view access (Figure 4). This may be explained by previous studies showing that "views of the outside for temporal information" is the strongest predictor of window size [5], and the latter can be considered a reliable proxy for access. Even though the connection to the outside has been measured using survey scales [24], this cannot be easily modeled in practice.

***Survey participants are limited to a specific group***

In this study, the research pipeline was directed toward the WV evaluation of residential space in urban context. However, the survey was distributed to a fairly uniform student group at an Architecture department, and this may prevent generalization of the results. Therefore, a broader extrapolation of this research to other populations (e.g. office-workers) and building types (e.g. offices, classrooms) may be needed to widen the utility of this tool. In response to this limitation, the authors provide the questionnaire and data as open-source resources [55] to enable future WV studies with a large and consistent dataset. Nevertheless, most participants have an architectural background and could have been more conscious about the design considerations of the environment they were asked to evaluate (i.e. their responses may not reflect the opinions of the lay population). Through future studies, responses collected from the cohort in this study could be compared with evauations collected from participants from a non-architectural background.

**4.4 Application of the research result**

This research builds a pipeline that can evaluate view satisfaction in the early design stage. To translate the utility of Seemo-Raycaster and Seemo-Predictor into practice, they are developed as a Grasshopper Plug-in embedded in Rhino3D. Designers can leverage the usage of Seemo tools by facilitating geotag information from OpenStreetMaps data through Urbano.io (Figure 3) and test an early architecture 3D model in Rhino 3D. Those 3D models can be easily linked with Seemo-Raycaster and Seemo-Predictor in Grasshopper, and the plug-ins compute view satisfaction prediction of the architecture model. Designers and planners can easily apply view evaluation in their design and simulation process to promote general (i.e. overall view) or certain (e.g. privacy) qualities of the window view. This application is not confined to architecture scale but can be utilized on an urban scale and contribute to zoning and real estate value assessment.

## 5. Conclusion

The authors developed a new view property computational tool, Seemo-Raycaster, that rapidly measures 23 WV variables postulated to influence view satisfaction. Based on Seemo-Raycaster's computation result and survey response, view satisfaction prediction tool, Seemo-Predictor, was developed. To verify the tool's validity and its suitability within the early stages of building design, its predictive performance was compared against an established WV framework [20], and to surveyed satisfaction data. The following objectives were answered from this exploration:

- Occupant's satisfaction data on various WVs consisting of different window configurations, and view content, was collected to expand understanding of WV quality and satisfaction.
- Trained supervised ML models were embedded in Seemo-Predictor to provide high prediction accuracies for overall view satisfaction, view content, and privacy, when view features measured by the Seemo-Raycaster acted as input data and matched with satisfaction survey responses.
- When comparing the predictions through two case studies, the tool gave closer estimates for view satisfaction than the framework. It is believed that the granularity and high fidelity of measurements offered by the tool facilitated more accurate predictions of satisfaction, advocating the need for detailed WV assessments.

The tool is capable of assisting in the design of residential WVs at the early building design stage. This not only meets the visual satisfaction requirements for residential occupants but may also help support their health and comfort needs in tandem. In consideration of the Covid-19 and the increased time people spend at their own residence, there is a growing need for tools to aid in the design of WVs, ensuring they remain connected to the outside.

## Acknowledgements

The authors would like to thank the Ministry of Land, Infrastructure, and Transportation of Korea as well as Korea Agency for Infrastructure Technology Advancement (KAIA) for funding this research through Architectural Design Human Resource Development Project.

# Appendices
## Appendix A: Ko's Framework

$$V_{content} = L_{sky} + L_{landscape} \cdot wf_{ct.dis.} + L_{ground} \cdot wf_{movement} + L_{nature} \cdot wf_{nature} \text{ with} \tag{A.1}$$

$$L_{sky}, L_{landscape}, L_{ground}, L_{nature} = \begin{cases} 0.25 \text{ if present in the scene} \\ 0 \text{ if absent in the scene} \end{cases} \tag{A.2}$$

$$wf_{ct.dis.} = \begin{cases} 1 \text{ if } 50\,m < content\ distance \\ 0.75 \text{ if } 20m < content\ distance \leq 50m \\ 0.5 \text{ if } 6m < content\ distance \leq 20m \\ 0 \text{ if } content\ distance \leq 6m \end{cases} \tag{A.3}$$

$$wf_{movement} = \begin{cases} 1 \text{ if distant dynamic feature(s)}(> 6m) \text{is present in the scene} \\ 0.5 \text{ if no dynamic featrue(s)}(\leq 6m) \text{is present in the scene} \\ 0 \text{ if nearby dynamic feature(s)}(\leq 6m) \text{is present in the scene} \end{cases} \tag{A.4}$$

$$wf_{nature} = \begin{cases} 1 \text{ if \% of natural reatures in the scene} > 50\% \\ 0.75 \text{ if } 25\% < \text{\% of natural features in the scene} \leq 50\% \\ 0.5 \text{ if \% of naatural features in the scene} \leq 25\% \\ 0 \text{ if no natural feature in the scene} \end{cases} \tag{A.5}$$

$$V_{access} = \begin{cases} 1 \text{ if } \alpha_{view} \geq \alpha_{saturation} \\ y \text{ if } \alpha_{min} < \alpha_{view} < \alpha_{saturation} \text{ with } y = \frac{1}{2}\left(\frac{\alpha_{view}}{\alpha_{saturation} - \alpha_{min}}\right) \\ 0.5 \text{ if } \alpha_{view} = \alpha_{min} \\ 0 \text{ if } \alpha_{view} < \alpha_{min} \end{cases} \tag{A.6}$$

**Table A7** View access reference angle

| $V_{content}$ | View content | $\alpha_{min}$ | $\alpha_{saturation}$ | Reference |
|---|---|---|---|---|
| 0.25 | Sky or ground view | Vertical view angle of 30° | - | (IWBS, 2019) |
| | Landscape view (no nature) | The smaller view angle of 11° | The smaller view angle of 90° | (Heschong Mahone Group, 2003) |
| 0.5 or higher | Landscape view with nature | The smaller view angle of 9° | The smaller view angle of 50° | (Heschong Mohone Group, 2003) |
| | Landscape view with sky or ground | Horizontal view angle of 14° | Horizontal view angle of 54° | (CEN/TC 169, 2019) |

# Appendix B: Case study 1 calculations for comparison

**Table B1** Comparison between Seemo-Raycaster, Seemo-Predictor, and the Framework

| | | 114 | | 234 | | 404 | | 424 | | 574 | | 584 | |
| --- | --- | --- | --- | --- | --- | --- | --- | --- | --- | --- | --- | --- | --- |
| | | 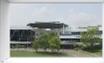 | | 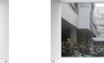 | | 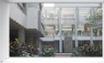 | | 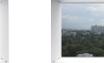 | | 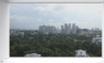 | | 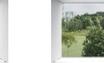 | |
| Sky View parameter (%) | Lsky(0-0.25) | 0.45 | 0.25 | 0.03 | 0.25 | 0.48 | 0.25 | 0.22 | 0.25 | 0.02 | 0 | 0 | 0 |
| Building View parameter (%) | Llandscape(0-0.25) | 0.19 | 0.25 | 0.67 | 0.25 | 0.16 | 0.25 | 0.01 | 0.25 | 0.23 | 0.25 | 0.67 | 0.25 |
| Distance to Building (m) | wfct.dist(0-1) | 87.13 | 1 | 13.8 | 0.5 | 123.02 | 1 | 354.05 | 1 | 33.67 | 1 | 5.47 | 0 |
| Equipment View parameter (%) | - | 0 | - | 0 | - | 0 | - | 0 | - | 0 | - | 0.31 | - |
| Distance to Equipment (m) | - | - | - | - | - | - | - | - | - | - | - | 5.09 | - |
| Tree View parameter (%) | Lnature(0-0.25) | 0.21 | 0.25 | 0.28 | 0.25 | 0.36 | 0.25 | 0.55 | 0.25 | 0.4 | 0.25 | 0 | 0 |
| Distance to Tree (m) | wfnature(0-1) | 52.43 | 0.75 | 8.04 | 0.5 | 171.18 | 1 | 15.24 | 1 | 2.73 | 1 | 0 | 0 |
| Pavement View parameter (%) | Lground(0-0.25) | 0.04 | 0.25 | 0.02 | 0.25 | 0 | 0.25 | 0 | 0.25 | 0 | 0.25 | 0.02 | 0 |
| Water View parameter (%) | - | 0 | - | 0 | - | 0 | - | 0 | - | 0.28 | - | 0 | - |
| Distance to Water (m) | - | - | - | - | - | - | - | - | - | 7.48 | - | - | - |
| Ground Vegetation View parameter(%) | - | 0.1 | - | 0 | - | 0 | - | 0.22 | - | 0.05 | - | 0 | - |
| Distance to Grass (m) | - | 55.25 | - | - | - | - | - | 26.32 | - | 23.76 | - | 0 | - |
| Dynamic View parameter (%) | wfmovement(0-1) | 0 | 0.5 | 0 | 0 | 0 | 0.5 | 0 | 0.5 | 0 | 1 | 0 | 0.5 |
| Distance to Dynamic (m) | - | - | - | - | - | - | - | - | - | - | - | - | - |
| View Content Prediction((-5)-5) | Vcontent(0-1) | 1.68 | 3.13 (0.813) | 1.94 | 0 (0.5) | 3.20 | 3.75 (0.875) | 2.67 | 3.75 (0.875) | 1.50 | 2.5 (0.75) | -4.11 | -5 (0) |

# Appendix C: Case study 2 calculations for comparison

**Table C1** Seemo-Predictor Comparison with the Survey, Framework, and Seemo-Raycaster

| | Image | 4 | 394 | 550 | 520 | 334 | 294 | 284 | 154 |
|---|---|---|---|---|---|---|---|---|---|
| | | 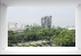 | 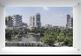 | 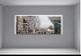 | 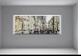 | 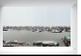 | 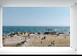 | 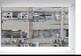 | 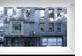 |
| Survey | **Overall Rating** $Z_1$ | 2.74 | 3.625 | -0.633 | 1.9666 | 3.1 | 2.6 | -2.925 | 0.1 |
| | **View Content** $Y_1$ | 2.38 | 3.3 | 0.8666 | 2.5 | 3.45 | 4.925 | -3.75 | -0.6 |
| | **View Access** $Y_2$ | 3.04 | 3.6 | 2.3 | 2.8 | 4.175 | 5 | 3.5 | 3.1666 |
| | **Privacy** $Y_3$ | 1.9 | 1.375 | -2.3 | -0.3 | 0.85 | -0.975 | -3.8 | -2.766 |
| Framework | VQI(-5~5) $Z_1$ Vcontent x Vaccess | -0.483 | -0.483 | -3.117 | -1.523 | -1.387 | -1.688 | -5 | -5 |
| | Vcontent(-5~5) $Y_1$ | 4.375 | 4.375 | 3.125 | 1.25 | 2.5 | 1.875 | -5 | -5 |
| | Vaccess(-5~5) $Y_{2,3}$ | -0.182 | -0.182 | -2.682 | 0.5625 | -0.182 | -0.182 | -2.5 | -2.5 |
| | Lsky(0-0.25) | 0.25 | 0.25 | 0.25 | 0.25 | 0.25 | 0.25 | 0 | 0 |
| | Llandscape(0-0.25) | 0.25 | 0.25 | 0.25 | 0.25 | 0.25 | 0 | 0.25 | 0.25 |
| | wfct.dist(0-1) | 1 | 1 | 0.75 | 0.5 | 1 | 0 | 0 | 0 |
| | Lnature(0-0.25) | 0.25 | 0.25 | 0.25 | 0 | 0 | 0.25 | 0 | 0 |
| | wfnature(0-1) | 0.75 | 0.75 | 0.5 | 0 | 0.5 | 0.75 | 0 | 0 |
| | Lground(0-0.25) | 0.25 | 0.25 | 0.25 | 0.25 | 0.25 | 0.25 | 0 | 0 |
| | wfmovement(0-1) | 1 | 1 | 1 | 1 | 1 | 1 | 0.5 | 0 |
| Seemo-Raycaster | Sky View parameter(%) | 0.42 | 0.33 | 0.16 | 0.03 | 0.49 | 0.51 | 0.03 | 0 |
| | Sky Condition | 2 | 2 | 1 | 1 | 1 | 2 | 0 | 0 |
| | Building View parameter(%) | 0.15 | 0.21 | 0.66 | 0.92 | 0.09 | 0 | 0.78 | 0.95 |
| | Distance to Building(m) | 138.94 | 704.41 | 43.92 | 17.52 | 152.68 | - | 10.32 | 7.59 |
| | Equipment View parameter(%) | 0 | 0 | 0 | 0 | 0 | 0 | 0.19 | 0 |
| | Distance to Equipment(m) | - | - | - | - | - | - | 10.32 | - |
| | Tree View parameter(%) | 0.36 | 0.32 | 0.03 | 0 | 0.02 | 0 | 0 | 0 |
| | Distance to Tree (m) | 52.25 | 327.69 | 88.27 | - | 734.73 | - | - | - |
| | Ground Vegetation View parameter(%) | 0 | 0.01 | 0.08 | 0 | 0 | 0.32 | 0 | 0 |
| | Distance to Ground Vegetation(m) | - | 641.48 | 21.01 | - | - | 47.7 | - | - |
| | Water View parameter(%) | 0 | 0.12 | 0 | 0 | 0.27 | 0.16 | 0 | 0 |
| | Distance to Water(m) | - | 319.62 | - | - | 214.31 | 161.05 | - | - |
| | Pavement View parameter(%) | 0.06 | 0 | 0.04 | 0 | 0.07 | 0 | 0 | 0 |
| | Dynamic View parameter(%) | 0.01 | 0.01 | 0.03 | 0.05 | 0.07 | 0.01 | 0 | 0.05 |
| | Distance to Dynamic(m) | 94.44 | 320.14 | 17.48 | 44.17 | 512.68 | 42.32 | - | 5.21 |
| Seemo-Predictor | **Overall Rating** $Z_1$ | 3.05 | 3.51 | 0.16 | 0.68 | 2.04 | 3.51 | -1.72 | -0.74 |
| | **View Content** $Y_1$ | 2.80 | 3.77 | 0.09 | 2.46 | 3.23 | 3.77 | -3.64 | -0.27 |
| | **View Access** $Y_2$ | 3.83 | 4.37 | 1.79 | 0.09 | 3.33 | 4.37 | 3.51 | 2.82 |
| | **Privacy** $Y_3$ | 1.73 | 2.29 | -0.87 | -0.12 | 2.13 | 2.29 | -1.42 | -1.42 |

# Appendix D: IRB Approval

Procedures outlined in this research were granted IRB approval (Protocol ID: 2011009973) according to Cornell IRB policy and under paragraph(s) 2 of the Department of Health and Human Services Code of Federal Regulations 45CFR 46.104(d).